\documentclass{article}

\usepackage[dblblindworkshop, final]{neurips_2026}
\usepackage{amsmath}
\usepackage{amssymb}
\usepackage[utf8]{inputenc}
\usepackage[T1]{fontenc}
\usepackage{hyperref}
\usepackage{url}
\usepackage{booktabs}
\usepackage{amsfonts}
\usepackage{nicefrac}
\usepackage{microtype}
\usepackage{xcolor}
\usepackage{graphicx}
\usepackage{algorithm}
\usepackage{algpseudocode}
\title{Match the Distribution, Not the Compute: Post-Training Multi-Token Prediction Heads}

\author{%
  Prachi Badarayani, Aidan Jay, Chenghui Zhou, Dayquan Julienne, \\\textbf{Yuan Gao, Tianwei Chen, George Zerveas, Ishmam Zabir, Xiren Zhou, Chris Quirk, Xia Song}\\
  Microsoft\\
}

\begin{document}

\maketitle

\begin{abstract}
Multi-token prediction (MTP) improves the throughput of autoregressive generation by enabling the language model to draft multiple next tokens per forward pass, while a verification step over draft tokens ensures that token distribution of the backbone is preserved. Every open MTP-family release (MiMo-7B, DeepSeek-V3, Qwen3) trains its heads jointly with the backbone over the full pretraining run of tens of trillions of tokens, thus setting the drafter quality at pretraining time. We ask whether a lightweight post-training pass on target-generated chain-of-thought is enough to reach the same expected throughput speedup on a frozen reasoning model, and study how a serving-time system built on such a checkpoint can be optimized. We present three findings. 1)~On a frozen Qwen3-8B with $K{=}3$ chained MTP heads, we show that a post-training recipe with plain cross-entropy on $\approx\!2.5$B tokens reaches or exceeds the expected speedup of jointly trained MiMo-7B on math, coding and knowledge benchmarks. Our post-training recipe utilizes $10^3$--$10^4\times$ less MTP-training tokens as compared with joint pre-training of MiMO-7B MTP baseline. 2)~We propose a chain-aware relaxation of draft token verification rule that allows a bounded drift from backbone language model token distribution. We show that this relaxation lifts expected speedups by $+12$ to $+16\%$ per benchmark while preserving task accuracy. 3)~We propose an adaptive controller that dynamically chooses the number of MTP heads to be engaged at inference time and demonstrate recovery of upto $11$--$14\%$ loss in speedup using fixed maximum MTP draft length. 
\end{abstract}

\section{Introduction}
Autoregressive decoding pays one forward pass per output token. Speculative decoding~\citep{leviathan2023spec,chen2023accelerating} amortises this by delegating $K$ candidate tokens to a lightweight drafter and verifying them in one target pass; the achieved throughput speedup tracks the mean accepted length (MAL) of the drafted tokens. Multi-token prediction (MTP) heads~\citep{gloeckle2024better,cai2024medusa,ankner2024hydra} predict $K$ next tokens from the target model's own hidden states. DeepSeek-V3~\citep{deepseekv3}, MiMo-7B~\citep{mimo2025} and EAGLE-3~\citep{li2025eagle3} train the MTP heads jointly with the backbone at the pretraining scale, which locks the drafter to a pretraining-time snapshot.

We ask whether the joint-pretraining cost is intrinsic to MTP quality or specific to the recipe. We show that with backbone-generated chain-of-thought as training data, post-trained MTP heads can match or outperform jointly pretrained MTP heads at $10^3$--$10^4\times$ less training compute. This enables us to explore post-trained MTP architectures and show improved throughput with Mamba chaining~\citep{gu2023mamba} of MTP heads, while continued pretraining of backbone is not helpful.

During MTP inference, classical speculative decoding enforces an exact distributional match that is stricter than task quality requires. We propose a chain-aware thresholding of the running joint probability of tokens to relax this distributional match requirement. We show that the expected throughput speedup as represented by Mean Accepted Length (MAL) consistently improves with this relaxation without compromising the task accuracy (Section~\ref{sec:approx_verify}). When MTP heads are used in production workloads, a fixed draft size reduces observed real throughput at higher batch sizes. Fixed draft sizes also prevent the ability to utilize longer MTP head predictions for predictable text generation at smaller batch sizes. Hence, we propose an adaptive controller that dynamically selects the number of MTP heads to be engaged by utilizing online throughput feedback and show improvements in MAL over defaulting to fixed draft length (Section~\ref{sec:adaptive}). 

\section{Post-training on target-generated chain-of-thought}
\label{sec:method}
An MTP head consumes hidden states $h_t = f_\theta(x_{<t})$ where $f_\theta$ is the $\theta$ parameterized backbone language model and $x_{<t}$ is a backbone-generated continuation on an input prompt. We have three MTP heads $g^1_{\theta_1}, g^2_{\theta_2}, g^3_{\theta_3}$ that produce the next token drafts $d_{t+1}, d_{t+2}$, and $d_{t+3}$ respectively, following $g^1_{\theta_1}(h_t, x_t)=d_{t+1}$, $g^2_{\theta_2}(h_t, d_{t+1})=d_{t+2}$ and $g^3_{\theta_3}(h_t, d_{t+2}) = d_{t+3}$. A distribution-preserving verification pass~\citep{leviathan2023spec} accepts a prefix of length $l \in \{0, 1,2,3\}$ of the draft tokens $[d_{t+1},.., d_{t+l}]$, thus producing $l+1$ tokens in one forward pass instead of single token produced by vanilla autoregressive model $f_\theta$. The expected throughput speedup or \emph{mean accepted length} (MAL) is $\mathbb{E}[l+1]$, the expected number of tokens committed per verification step: with per-head acceptance rates $\alpha_1,\ldots,\alpha_K$, $\mathrm{MAL} = 1 + \sum_{k=1}^{K}\prod_{j=1}^{k}\alpha_j$.

\paragraph{Architecture and objective.} We adopt the chained MTP head of \citet{mimo2025}: head $g^k_{\theta_k}$ takes as input the concatenation of $h_t$ with the argmax  of unembedding of head $g^{k-1}_{\theta_{k-1}}$, i.e., token $d_{k}$. The architecture of each head is one transformer layer at the backbone's width, that internally embeds the input token and also shares the frozen language-model's unembedding head. Each head adds $\approx\!200$M trainable parameters, matching MiMo-7B's per-head budget. Training minimises per-head cross-entropy $\mathcal L = -\tfrac{1}{K}\sum_k\sum_t \log p^{(k)}(d_{t+k}\mid h_t, d_{t+k-1})$ where $p^{(k)}$ is the probability given by MTP head $g^k_{\theta_k}$ and when $k=0$, $d_t=x_t$. Only head parameters $\theta_k$ are updated while backbone parameters $\theta$ are frozen. Head $k$ conditions on the argmax of head $k{-}1$ (i.e., $d_{k}$) at training, matching the inference-time chain. We train $K{=}3$ MTP heads for the main empirical results and a $K{=}6$ variant for the adaptive MTP controller  Section~\ref{sec:adaptive}.

\paragraph{On-policy target-generated corpus.} We use $1.89$M prompts from AM-Thinking~\citep{amthinking2024} to train the MTP heads; it spans general chat, code, math, and instruction following. We sample a complete chain-of-thought response from Qwen3-8B in thinking mode for each prompt (up to $8$K tokens per response), and pack prompt-response pairs into fixed-length training sequences of $2048$ tokens, yielding $\approx\!2.5$B training tokens per epoch. Appendix~\ref{app:reproducibility} provides full setup details. The closest prior recipes are DistillSpec~\citep{zhou2024distillspec} and EAGLE-3~\citep{li2025eagle3}.

\section{Chain-aware approximate verification}
\label{sec:approx_verify}
Classical speculative decoding accepts a drafted token $ d_{t+k}$ if a uniform draw is below $p^{(k)}(d_{t+k})/f_\theta(d_{t+k})$ with rejection sampling, preserving the target's output distribution exactly~\citep{leviathan2023spec}. For reasoning targets whose distribution is peaked this guarantee is stronger than what task quality actually needs. Prior relaxations trade a small distributional slack for higher MAL by evaluating each draft position in isolation (typical acceptance~\citep{cai2024medusa,meister2022typical}, top-$k$ agreement, BiLD divergence gates~\citep{kim2023bild}, and the continuous knobs of DistillSpec~\citep{zhou2024distillspec}, Cactus~\citep{hao2026cactus} and Cascades~\citep{narasimhan2025cascades}). These per-position criteria discard a signal that the chain of $K$ draft positions carries jointly: a single low-probability position in an otherwise confident chain is different from two consecutive mediocre positions. We propose a chain-aware relaxation of the strict distributional match guarantee, that accepts the maximal draft prefix $d_{t+1:m^{*}}$ of length $m^{*}$ such that its running geometric-mean target probability stays above a threshold $\tau$:
\begin{equation}
m^\star = \max\!\Big\{m\le K : \Big(\textstyle\prod_{k=1}^{m} p^{(k)}(d_{t+k})\Big)^{1/m} \!\ge\! \tau\Big\}.
\label{eq:joint}
\end{equation}

\section{Adaptive draft depth}
\label{sec:adaptive}
Increasing the number of MTP heads increases the expected throughput speedup (MAL) monotonically, but observed runtime throughput speedup usually peaks at three MTP heads due to GPU memory bottleneck~\citep{deepseekv3}. We observe that at low batch-size or highly predictable texts, a higher number of MTP heads amortise the verification cost against a single-request forward pass and high number of MTP heads $K$ can pay off. However, at high batch-size or hard-to-predict text sequences, the drafter becomes the bottleneck as deeper MTP heads no longer cover the extra verification costs, and low $K$ provides highest runtime speedup. Production observations~\citep{nightjar2025} confirm speculation is often not worth running above a batch threshold, so practical recommendations dictate switching off MTP or using fixed 3 MTP heads to avoid throughput degradation.

We propose an adaptive controller that dynamically selects the right number of MTP heads $K$ at runtime such that optimal throughput can be achieved under variable batch size and text prediction difficulty. We use our post-training recipe (Section \ref{sec:method}) to learn $K{=}6$ MTP heads. We implement an online lower confidence bound bandit controller~\citep{auer2002ucb,lattimore2020bandits} with 6 arms that selects MTP draft depth $\{K{=}1,\ldots,K{=}6\}$. Each decoding window credits the active arm with observed batched throughput, updating a per-arm running mean. After a short round-robin exploration of all 6 arms, we exploit the running argmax with periodic re-exploration and \emph{commit} once one arm leads for $C$ windows by a relative margin $\delta$; if no arm commits within $W_{\max}$ windows we fall back to the arm with the highest lower-confidence bound (mean minus one standard deviation). Independent throughput samples make each running mean concentrate on its true mean at rate $O(1/\sqrt{N_a})$ where $N_a$ is number of samples observed for arm $a$. Gaps below $\delta$ can commit to a suboptimal arm (loss $\le\delta$). You can find the hyperparameters and pseudocode in Appendix~\ref{app:reproducibility}.

\section{Experiments}
\label{sec:results}

\paragraph{Setup.}
We evaluate on four math reasoning benchmarks in thinking mode and five benchmarks on other areas like coding, knowledge, etc. Backbone: Frozen Qwen3-8B~\citep{qwen3} with $K{=}3$ chained MTP heads. Baseline: MiMo-7B-RL~\citep{mimo2025} with per-head architecture and parameter count matched to ours (App.~\ref{app:baseline_choice}). Both systems run exact rejection-sampling verification at batch $4$ on a single A100-80GB GPU in vLLM. 

\begin{table}[!ht]
\centering
\small
\caption{Mean accepted length and per-head acceptance rates ($H_1/H_2/H_3$, \%) at matched per-head capacity. Our MAL generally matches or exceeds MiMo on all benchmarks. Task accuracy is preserved by exact verification.}
\label{tab:headline}
\setlength{\tabcolsep}{5pt}
\begin{tabular}{lccccc}
\toprule
& \multicolumn{2}{c}{MAL} & & \multicolumn{2}{c}{$H_1/H_2/H_3$ (\%)} \\
\cmidrule(lr){2-3}\cmidrule(lr){5-6}
Benchmark & Ours & MiMo-7B & Ratio & Ours & MiMo-7B \\
\midrule
\multicolumn{6}{l}{\textit{Math reasoning}} \\
GSM8K~\citep{gsm8k}    & $\mathbf{3.05}$ & $2.80$          & $1.09\times$ & $86.5/67.4/51.1$ & $87.1/59.1/33.6$ \\
MATH500~\citep{lightman2023lets} & $\mathbf{3.00}$ & $2.87$          & $1.05\times$ & $86.1/65.6/48.3$ & $89.1/62.4/35.3$ \\
AIME24~\citep{aime2024}  & $2.81$          & $\mathbf{2.85}$ & $0.99\times$ & $82.9/58.5/39.4$ & $88.9/61.7/34.3$ \\
AIME25~\citep{aime2025}  & $\mathbf{2.81}$ & $2.82$          & $1.00\times$ & $83.1/58.6/39.3$ & $88.3/60.6/33.1$ \\
\midrule
\multicolumn{6}{l}{\textit{knowledge / code / open-ended }} \\
MMLU-Redux~\citep{gema2024mmluredux} & $\mathbf{2.64}$ & $2.54$ & $1.04\times$ & $79.2/52.5/32.7$ & $80.5/48.7/24.4$ \\
MBPP~\citep{austin2021mbpp}    & $\mathbf{2.84}$ & $2.70$ & $1.05\times$ & $84.0/59.3/40.9$ & $85.9/55.2/28.6$ \\
MT-Bench~\citep{zheng2023mtbench} & $\mathbf{2.67}$ & $2.40$ & $1.11\times$ & $78.8/53.2/35.1$ & $76.8/43.1/20.2$ \\
Alpaca~\citep{taori2023alpaca}  & $\mathbf{2.47}$ & $2.23$ & $1.11\times$ & $74.4/45.9/26.8$ & $72.0/36.0/14.9$ \\
CNN/DailyMail~\citep{nallapati2016cnndm}  & $\mathbf{2.32}$ & $2.24$ & $1.04\times$ & $70.9/40.1/21.1$ & $72.3/35.8/15.6$ \\
\bottomrule
\end{tabular}
\end{table}

\paragraph{Post-training matches or exceeds joint pretraining on MAL.}
On MAL, Post-trained heads generally reach or exceed MiMo-7B on all benchmarks. First-head acceptance slightly favours MiMo ($\sim\!87$--$89\%$ vs.\ ours $\sim\!83$--$87\%$ on math benchmarks) as joint pretraining gives a small edge to first position. The picture inverts on the deepest head: MiMo saturates around $33$--$35\%$ on every math benchmark, while our third head reaches higher acceptance probabilities despite identical MTP chaining architecture. Ablations over chaining MTP heads, continued pretraining of backbone and training loss functions done in Appendix~\ref{app:ablations} show that Mamba chaining~\citep{gu2023mamba} of MTP head improves MAL by exploiting draft token dependence, while switching to KL training loss w.r.t. backbone or performing continued pretraining of the backbone model does not help.  

\paragraph{Approximate verification.}
We compare Eq.~\ref{eq:joint} against five verification relaxation baselines, each parameterised by a single knob: \emph{argmax bypass}, \emph{typical acceptance}~\citep{cai2024medusa,meister2022typical}, \emph{BiLD}~\citep{kim2023bild}, \emph{DistillSpec} $\beta$~\citep{zhou2024distillspec}, \emph{Cactus} $\lambda$~\citep{hao2026cactus}, and \emph{Cascades} $\tau_c$~\citep{narasimhan2025cascades}. For every rule we sweep its knob and report the point that maximises MAL per benchmark subject to task accuracy staying within $\pm 3$~pp of exact verification (each column of Table~\ref{tab:verify} is therefore accuracy-preserving by construction; cells outside the corridor at every parameter we tested are marked \emph{unsafe}). On GSM8K, MATH500 and AIME 2024 the top methods (BiLD and the chain-aware rule) are within $0.02$--$0.06$ MAL of each other. The four rules separate on AIME 2025: BiLD leaves the accuracy corridor at every threshold we swept, while the chain-aware rule stays inside it at $\tau{=}0.30$ (MAL $3.17$, accuracy $73.3\%$ vs.\ exact $76.7\%$). Averaged over the four benchmarks the chain-aware rule reaches $3.31$, a $+14\%$ MAL lift over exact verification at accuracy preserved on every benchmark.

\begin{table}[!ht]
\centering\small
\caption{Chain-aware approximate verification shows greatest improvement in expected throughput speedup (MAL). For all cells with valid entry, the task accuracy was preserved despite lenient verification.}
\label{tab:verify}
\setlength{\tabcolsep}{4pt}
\begin{tabular}{lccccc}
\toprule
Rule (deterministic $Q{=}1$)                                 & GSM8K & MATH500 & AIME24 & AIME25 & Avg \\
\midrule
Standard (exact)                                             & $3.05$ & $3.00$ & $2.81$ & $2.81$ & $2.91$ \\
Argmax bypass                                                & $3.13$ & $3.09$ & $2.86$ & $2.85$ & $2.98$ \\
Typical acceptance \citep{cai2024medusa,meister2022typical}  & $3.40$ & $3.34$ & $3.14$ & $\textbf{3.19}$ & $3.27$ \\
BiLD KL threshold \citep{kim2023bild}                        & $3.42$ & $3.34$ & $3.11$ & \emph{unsafe} & --- \\
DistillSpec $\beta$ \citep{zhou2024distillspec}              & $3.24$ & $3.19$ & $3.00$ & $3.01$ & $3.11$ \\
Cactus $\lambda$ \citep{hao2026cactus}                       & $3.33$ & $3.24$ & $3.15$ & $2.99$ & $3.18$ \\
Cascades $\tau_c$ \citep{narasimhan2025cascades}             & $3.10$ & $3.04$ & $2.87$ & $2.81$ & $2.96$ \\
\midrule
\textbf{Chain-aware} (Eq.~\ref{eq:joint}, best-safe $\tau$ per bench) & $\mathbf{3.43}$ & $\mathbf{3.38}$ & $\mathbf{3.25}$ & $3.17$ & $\mathbf{3.31}$ \\
\bottomrule
\end{tabular}
\end{table}
\paragraph{Adaptive Multi-Token Prediction.}
Table~\ref{tab:adaptive} summarises the benefit of adaptive controller averaged across ten workloads (per-workload breakdown in Table~\ref{tab:adaptive_bandit}, Appendix~\ref{app:adaptive_workshop}). On every workload where deep MTP heads help, the adaptive controller chooses $K=6$; on the ones where it hurts, the controller falls back to smaller $K$ values. Fixed $K{=}6$ forfeits $11$--$14\%$ throughput at batches $64$/$128$ as deep-head verification overtakes acceptance gain. The adaptive MTP controller removes the trade while paying an overhead cost: it captures the deep-head upside where $K{=}6$ helps and falls back to a low $K$ when batch size grows and GPU contention rises. The bandit algorithm overhead loses up to 6\% on speedup ratio on an average at low batch size, but recovers most of the lost throughput at high batch sizes that are common in production.  

\begin{table}[!ht]
\centering
\small
\caption{Adaptive controller selects the right MTP depth dynamically for a workload. Table shows averaged speedup ratio w.r.t fixed MTP depth of $K=3$ across ten workloads.}
\label{tab:adaptive}
\setlength{\tabcolsep}{6pt}
\begin{tabular}{lcccc}
\toprule
Observed throughput ratio to fixed $K{=}3$ (avg over 10 workloads) & bs $4$ & bs $16$ & bs $64$ & bs $128$ \\
\midrule
Fixed $K{=}6$              & $1.01$ & $1.01$ & $0.86$ & $0.89$ \\
Adaptive-$K$ controller    & $0.94$ & $0.94$ & $\mathbf{0.98}$ & $\mathbf{0.99}$ \\
\bottomrule
\end{tabular}
\end{table}

\paragraph{Discussion and Conclusion.} Post-training on target-generated chain-of-thought matches or exceeds a jointly pretrained MTP baseline at $10^3$--$10^4\times$ less MTP-training compute on math, coding, instruction-following and knowledge benchmarks. Post-training also allows exploration of architecture choices that pretraining forecloses: a Mamba-style chaining of MTP heads improves expected speedups, while full-vocabulary KL loss function and continued pretraining of backbone do not improve the expected speedups(Table~\ref{tab:ablations}). Self-distillation of training data produces the highest gains in expected speedups. Exact rejection sampling is unnecessarily conservative on reasoning targets; our chain-aware relaxation shows consistent gains in expected throughput speedup (MAL) while maintaining accuracy of the generated outcomes. Online adaptive MTP controller allows us to dynamically choose the right number of MTP heads in response to changing batch sizes, GPU contention and predictability of generated text. 

\bibliographystyle{plainnat}
\bibliography{references}

\newpage
\appendix

\section{Ablations}
\label{app:ablations}
Table~\ref{tab:ablations} disentangles two comparisons. \emph{Row~1 vs.\ Row~2} isolates the effect of self-distillation: with the same corpus size and the same recipe otherwise, training on target-generated CoT instead of raw AM-Thinking text gains $+0.14$ to $+0.41$ MAL. \emph{The lower block} reports extensions trained on top of the same raw AM-Thinking baseline (Row~2), so they should be compared against Row~2 rather than the self-distilled base. A GRU~\citep{cho2014gru} cross-head gate is a small loss on the same-corpus baseline; a Mamba~\citep{gu2023mamba} gate shows gain of $+0.28$ MAL on average (up to $+0.43$ on GSM8K) but still falls short of self-distillation on three of four math benchmarks. LoRA~\citep{hu2021lora} adapters on the backbone are neutral to negative in both shallow ($4$-layer) and deep ($18$-layer) configurations. Full-vocabulary KL against the backbone's next-token distribution does not improve MAL. The pattern is consistent with the alignment argument of Section~\ref{sec:method}: the on-policy target-generated corpus already removes the covariate shift these interventions try to correct. Concrete architectures and hyperparameters are in Section~\ref{app:reproducibility}.

\begin{table}[!ht]
\centering
\small
\caption{Ablations. MAL on math reasoning tasks shows most improvement with cross-entropy loss and using self-distilled tokens from Qwen3-8B model. We set $K{=}3$ chained heads, exact verification, batch size $4$. Row~1 vs.\ Row~2 isolates self-distillation; the lower block is extensions trained on the same raw AM-Thinking corpus as Row~2 and should be compared against Row~2, not against the self-distilled base. Mamba gate chaining produces the most improvement in ablation settings over Row~2.}
\label{tab:ablations}
\setlength{\tabcolsep}{4pt}
\begin{tabular}{llcccc}
\toprule
Variant & Change & GSM8K & MATH500 & AIME24 & AIME25 \\
\midrule
\multicolumn{6}{l}{\emph{Effect of self-distillation}} \\
Base recipe        & Self-distilled CoT $+$ frozen $+$ CE & $\mathbf{3.05}$ & $\mathbf{3.00}$ & $\mathbf{2.78}$ & $\mathbf{2.81}$ \\
Raw AM-Thinking    & Same recipe, external text          & $2.64$ & $2.67$ & $2.59$ & $2.57$ \\
\midrule
\multicolumn{6}{l}{\emph{Extensions trained on Row~2's raw corpus}} \\
$+$ GRU gate       & Cross-head routing                  & $2.72$ & $2.50$ & $2.48$ & $2.46$ \\
$+$ Mamba gate     & Cross-head routing                  & $3.07$ & $2.97$ & $2.70$ & $2.71$ \\
$+$ Shallow LoRA   & Unfreeze 4 latest backbone layers      & $2.35$ & $2.32$ & $2.20$ & $2.20$ \\
$+$ Deep LoRA      & Unfreeze 18 latest backbone layers        & $2.59$ & $2.67$ & $2.50$ & $2.46$ \\
$+$ Full-vocab KL  & Match softmax not tokens            & $2.40$ & $2.31$ & $2.16$ & $2.14$ \\
\bottomrule
\end{tabular}
\end{table}

\section{Coverage outside math datasets}
\label{app:ood_workshop}
Our post-trained heads exceed MiMo-7B on MAL for MMLU-Redux~\citep{gema2024mmluredux}, MBPP~\citep{austin2021mbpp}, MT-Bench~\citep{zheng2023mtbench} (chat, $n{=}80$), Alpaca~\citep{taori2023alpaca} (instruction-following, $n{=}800$), and CNN/DailyMail~\citep{nallapati2016cnndm} (summarisation, $n{=}500$). All numbers are listed in Table~\ref{tab:headline}; the deep-head gap widens sharply on MT-Bench ($+15$pp on $H_3$) and Alpaca ($+12$pp), consistent with the reasoning-only headline: on-policy target-generated CoT trains a chain-aware drafter whose margin against joint pretraining is preserved outside the training domain. Task-accuracy scoring on MT-Bench and Alpaca requires an LLM-as-judge, which we do not commit to here; on CNN/DailyMail we report ROUGE against the reference highlights in \S\ref{app:tau_workshop}.

\section{Full $\tau$ sweep for chain-aware verification}
\label{app:tau_workshop}
We sweep the chain-aware $\tau$ of Eq.~\ref{eq:joint} across eight decades from $\tau{=}10^{-15}$ (accept-everything ceiling, verified against the internal $\log$-space clamp at $\log 10^{-12}\!=\!-27.6$) to $\tau{=}0.90$ (near-strict) on the four math benchmarks (Table~\ref{tab:tau_math}) and on CNN/DailyMail (Table~\ref{tab:cnndm_tau}). Two effects are worth noting.

\paragraph{MAL generally increases as $\tau$ decreases.} As $\tau$ is relaxed, more and more drafts are naturally accepted despite lack of perfect distribution match with the backbone. However, at $\tau{=}10^{-6}$ the head-1 gate is loose ($H_1{=}95\%$) but the geometric-mean penalty compounds and rejects half of the head-2 positions ($H_2{=}49$--$55\%$), giving a lower overall MAL than at $\tau{=}10^{-3}$ where the head-1 gate is tighter but the chain does not compound as harshly. Below $\tau{=}10^{-12}$ the log-space clamp makes the chain criterion trivially satisfied and MAL rises to the $K{+}1{=}4$ ceiling.

\begin{figure}[!ht]
\centering
\includegraphics[width=0.7\textwidth]{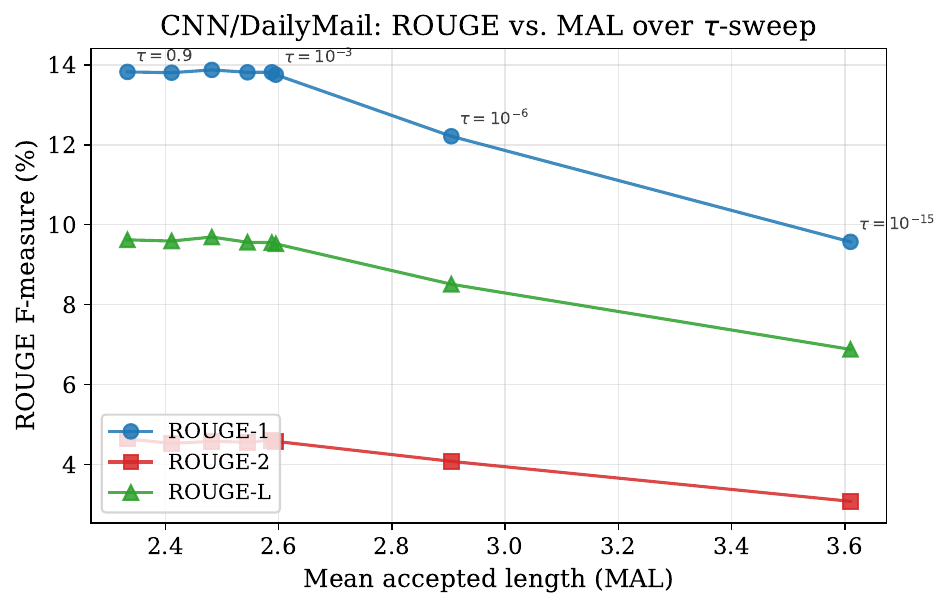}
\caption{ROUGE vs.\ MAL on CNN/DailyMail as $\tau$ varies over the eight decades in Table~\ref{tab:cnndm_tau}. Curves are nearly flat across the safe corridor and drop toward the ceiling. Contrast with the step-function behaviour of math accuracy in Table~\ref{tab:tau_math}: $n$-gram overlap tolerates local drift; exact final-answer grading does not.}
\label{fig:tau_pareto_workshop}
\end{figure}

\paragraph{Math accuracy is a step function of MAL; ROUGE is a smooth curve.} On math the safe corridor $\tau \in [10^{-3},\,0.90]$ preserves task accuracy at the exact-verification baseline; once $\tau$ leaves the corridor, AIME24/25 collapse to $0\%$ and MATH500 drops to $32\%$ at the ceiling. Math grading is brittle: a single wrong numerical step corrupts the \texttt{\textbackslash boxed\{$\cdot$\}} answer. On CNN/DailyMail the same $\tau$ sweep gives a smooth ROUGE curve (Fig.~\ref{fig:tau_pareto_workshop}): ROUGE-1/L stays flat at $13.75$--$13.87$ across the corridor and falls to $9.57$/$6.88$ at the ceiling because $n$-gram overlap rewards local topical fluency, which the base model retains even under corrupted decoding. This is a property of the metric, not of the recipe. For any deployment graded on final-answer correctness the operating range is $\tau \in [10^{-3},\,0.90]$, and the peak is at $\tau{=}10^{-3}$ ($+0.30$ to $+0.45$ MAL over exact verification).

\begin{table}[!ht]
\centering\small
\caption{Progressive-joint $\tau$ sweep on the four math benchmarks (K$=$3 checkpoint, batch $4$, thinking mode). $\tau{=}10^{-15}$ is the accept-everything ceiling; higher $\tau$ approaches strict verification. Task accuracy stays at the exact baseline for $\tau \geq 10^{-3}$ and collapses below; MAL is non-monotone in $\tau$ (see text).}
\label{tab:tau_math}
\setlength{\tabcolsep}{3pt}\scriptsize
\begin{tabular}{lcccccccc}
\toprule
$\tau$ & \multicolumn{2}{c}{GSM8K ($n{=}200$)} & \multicolumn{2}{c}{MATH500 ($n{=}100$)} & \multicolumn{2}{c}{AIME24 ($n{=}30$)} & \multicolumn{2}{c}{AIME25 ($n{=}30$)} \\
\cmidrule(lr){2-3} \cmidrule(lr){4-5} \cmidrule(lr){6-7} \cmidrule(lr){8-9}
       & MAL & acc\% & MAL & acc\% & MAL & acc\% & MAL & acc\% \\
\midrule
$10^{-15}$ (ceiling) & $3.85$ & $60.5$ & $3.93$ & $32.0$ & $3.86$ & $0.0$ & $3.89$ & $0.0$ \\
$10^{-6}$            & $3.44$ & $90.5$ & $3.07$ & $67.0$ & $2.94$ & $23.3$ & $2.98$ & $16.7$ \\
$10^{-3}$            & $\mathbf{3.43}$ & $97.0$ & $\mathbf{3.38}$ & $81.0$ & $\mathbf{3.25}$ & $80.0$ & $\mathbf{3.20}$ & $60.0$ \\
$10^{-2}$            & $3.43$ & $96.0$ & $3.36$ & $79.0$ & $3.14$ & $66.7$ & $3.18$ & $63.3$ \\
$0.10$               & $3.42$ & $96.5$ & $3.37$ & $80.0$ & $3.10$ & $63.3$ & $3.17$ & $63.3$ \\
$0.30$               & $3.33$ & $97.5$ & $3.30$ & $81.0$ & $3.14$ & $56.7$ & $3.17$ & $73.3$ \\
$0.50$               & $3.21$ & $96.0$ & $3.19$ & $80.0$ & $2.92$ & $70.0$ & $2.96$ & $70.0$ \\
$0.90$               & $3.08$ & $96.0$ & $3.06$ & $80.0$ & $2.81$ & $70.0$ & $2.81$ & $66.7$ \\
Exact (baseline)     & $3.05$ & $96.0$ & $3.00$ & $79.0$ & $2.81$ & $76.7$ & $2.81$ & $76.7$ \\
\bottomrule
\end{tabular}
\end{table}

\begin{table}[!ht]
\centering\small
\caption{$\tau$ sweep on CNN/DailyMail ($n{=}500$, K$=$3, thinking) with ROUGE-1/2/L (\%) against reference highlights. ROUGE tracks quality smoothly, unlike math accuracy.}
\label{tab:cnndm_tau}
\setlength{\tabcolsep}{6pt}
\begin{tabular}{lccccc}
\toprule
$\tau$ & MAL & R-1 & R-2 & R-L & tok/s \\
\midrule
$10^{-15}$ (ceiling) & $3.61$ & $9.57$ & $3.08$ & $6.88$ & $245$ \\
$10^{-6}$            & $2.90$ & $12.21$ & $4.08$ & $8.51$ & $208$ \\
$10^{-3}$            & $2.60$ & $13.75$ & $4.58$ & $9.52$ & $207$ \\
$10^{-2}$            & $2.59$ & $13.81$ & $4.61$ & $9.55$ & $208$ \\
$0.10$               & $2.54$ & $13.81$ & $4.56$ & $9.56$ & $204$ \\
$0.30$               & $2.48$ & $13.87$ & $4.58$ & $9.69$ & $197$ \\
$0.50$               & $2.41$ & $13.80$ & $4.53$ & $9.59$ & $195$ \\
$0.90$               & $2.33$ & $13.82$ & $4.64$ & $9.62$ & $187$ \\
Exact (baseline)     & $2.32$ & $13.75$ & $4.58$ & $9.55$ & $224$ \\
\bottomrule
\end{tabular}
\end{table}

\section{Adaptive draft depth: multi-benchmark coverage}
\label{app:adaptive_workshop}
Table~\ref{tab:adaptive} in the main text characterises the bandit on GSM8K. Table~\ref{tab:adaptive_bs_summary} shows the multi-workload picture, which additionally includes HumanEval~\citep{chen2021humaneval} and MMLU-Pro~\citep{wang2024mmlupro} beyond the datasets of Table~\ref{tab:headline}. Committing to fixed $K{=}6$ chases the low-batch MAL upside (matching $K{=}3$ within noise at low batch, $+1\%$) but collapses at high batch by $11$--$14\%$ because the deep heads' verification cost overtakes their acceptance gain. The online bandit sits at $93$--$98\%$ of fixed $K{=}3$ everywhere and $\geq\!97\%$ at high batch by falling back correctly. The bandit is therefore a safe replacement for fixed $K{=}3$ (worst cell $0.90$) and a strict Pareto improvement over fixed $K{=}6$ (which loses up to $29\%$). Table~\ref{tab:adaptive_bandit} extends this to per-workload cells and Table~\ref{tab:adaptive_coverage} documents the underlying $K{=}6$ head profile.

\begin{table}[!ht]
\centering
\caption{Bandit vs.\ fixed-$K$ throughput averaged across ten workloads (paired per-sample-median tok/s). Fixed $K{=}6$ crashes at high batch; the bandit reaches $\geq\!97\%$ of fixed $K{=}3$ there by falling back correctly.}
\label{tab:adaptive_bs_summary}
\setlength{\tabcolsep}{6pt}\small
\begin{tabular}{lcccc}
\toprule
Arm ratio to fixed $K{=}3$ & bs $4$ & bs $16$ & bs $64$ & bs $128$ \\
\midrule
Fixed $K{=}6$              & $1.01$ & $1.01$ & $0.86$ & $0.89$ \\
Adaptive bandit (online)   & $0.93$ & $0.94$ & $\mathbf{0.98}$ & $\mathbf{0.99}$ \\
\bottomrule
\end{tabular}
\end{table}

\begin{table}[!ht]
\centering\small
\caption{Per-cell view. Cell format `K=6 lift over K=3 \textbar\ bandit \% of best-fixed'. K=6 helps at low batch on math/code and hurts at high batch by $20$--$29\%$ on math/MCQ; the bandit stays close to whichever fixed is best. Mixes: \emph{Math (mixed difficulty)} draws $120/96/20/20$ from GSM8K/MATH500/AIME 2024/2025; \emph{Cross-domain} draws $96/80/80$ from GSM8K/MBPP/MMLU-Redux.}
\label{tab:adaptive_bandit}
\setlength{\tabcolsep}{3pt}
\begin{tabular}{lcccc}
\toprule
Workload & bs $4$ & bs $16$ & bs $64$ & bs $128$ \\
\midrule
GSM8K              & $+2$\%\,\textbar\,$90$\%  & $+3$\%\,\textbar\,$91$\%  & $-23$\%\,\textbar\,$96$\%  & $-21$\%\,\textbar\,$102$\% \\
MATH500            & $+3$\%\,\textbar\,$89$\%  & $+4$\%\,\textbar\,$89$\%  & $-22$\%\,\textbar\,$96$\%  & $-23$\%\,\textbar\,$98$\% \\
AIME 2024          & $-5$\%\,\textbar\,$90$\%  & $-3$\%\,\textbar\,$91$\%  & $-1$\%\,\textbar\,$96$\%   & $-2$\%\,\textbar\,$96$\% \\
AIME 2025          & $+1$\%\,\textbar\,$92$\%  & $+1$\%\,\textbar\,$93$\%  & $+1$\%\,\textbar\,$97$\%   & $-0$\%\,\textbar\,$98$\% \\
HumanEval          & $+9$\%\,\textbar\,$86$\%  & $+5$\%\,\textbar\,$92$\%  & $-4$\%\,\textbar\,$104$\%  & $+35$\%\,\textbar\,$74$\% \\
MBPP               & $-0$\%\,\textbar\,$96$\%  & $+0$\%\,\textbar\,$96$\%  & $-8$\%\,\textbar\,$102$\%  & $-4$\%\,\textbar\,$94$\% \\
MMLU-Pro           & $-1$\%\,\textbar\,$92$\%  & $+2$\%\,\textbar\,$93$\%  & $-21$\%\,\textbar\,$96$\%  & $-20$\%\,\textbar\,$99$\% \\
MMLU-Redux         & $-4$\%\,\textbar\,$95$\%  & $-7$\%\,\textbar\,$93$\%  & $-27$\%\,\textbar\,$97$\%  & $-29$\%\,\textbar\,$98$\% \\
\emph{Math (mixed difficulty)} & $+2$\%\,\textbar\,$89$\%  & $+4$\%\,\textbar\,$89$\%  & $-21$\%\,\textbar\,$96$\%  & $-22$\%\,\textbar\,$98$\% \\
\emph{Cross-domain mix}    & $+1$\%\,\textbar\,$93$\%  & $+0$\%\,\textbar\,$92$\%  & $-17$\%\,\textbar\,$96$\%  & $-20$\%\,\textbar\,$101$\% \\
\bottomrule
\end{tabular}
\end{table}

\begin{table}[!ht]
\centering\small
\caption{K$=$6 checkpoint per-head acceptance ($H_1$--$H_6$, \%) and MAL at batch $4$ across ten workloads (arm \texttt{fixed\_k6} of the bandit sweep).}
\label{tab:adaptive_coverage}
\setlength{\tabcolsep}{3pt}
\begin{tabular}{lcc}
\toprule
Workload & MAL & $H_1/H_2/H_3/H_4/H_5/H_6$ (\%) \\
\midrule
GSM8K              & $3.91$ & $91/78/65/31/18/10$ \\
MATH500            & $4.05$ & $92/80/68/35/21/13$ \\
AIME 2024          & $3.60$ & $89/72/56/24/13/7$  \\
AIME 2025          & $3.68$ & $89/73/57/27/14/8$  \\
\emph{Math (mixed difficulty)} & $3.94$ & $91/78/64/32/18/10$ \\
HumanEval          & $3.90$ & $90/74/58/36/20/12$ \\
MBPP               & $3.50$ & $85/68/50/30/14/8$  \\
MMLU-Redux         & $2.96$ & $80/57/37/14/6/2$   \\
MMLU-Pro           & $3.56$ & $86/68/52/26/15/9$  \\
\emph{Cross-domain mix} & $3.38$ & $85/66/48/22/11/6$ \\
\bottomrule
\end{tabular}
\end{table}

\section{Baseline choice: MiMo-7B-RL and Qwen3-8B}
\label{app:baseline_choice}
We report against MiMo-7B-RL~\citep{mimo2025} as the only publicly available jointly-pretrained MTP baseline that matches our reporting protocol. The alternatives fail this in at least one of three concrete ways: (i) DeepSeek-V3~\citep{deepseekv3} releases a single MTP head, not the $K{=}3$ chain we need to compare per-head profiles; (ii) EAGLE-3~\citep{li2025eagle3} uses a single recurrent draft module, not per-depth heads, so per-head acceptance is not defined in the same sense; (iii) the largest open MTP-family checkpoints are attached to backbones an order of magnitude larger than our target, so any MAL comparison would confound drafter quality with base-model quality. MiMo-7B-RL releases all three chained heads on an RL-tuned backbone at $\approx\!7$B parameters, which is the smallest gap to the natural target size for a $\approx\!200$M-per-head chain of ours.

We choose Qwen3-8B~\citep{qwen3} as the target because it is the closest freely available base model to MiMo-7B in parameter count (within $\approx\!15\%$) that reports competitive quality on the four math benchmarks used in Table~\ref{tab:headline}. A $70$B-class target would change every downstream number and confound the training-recipe question we ask: with the same base-model quality budget, how close to joint pretraining does a plain post-training pass on the chained heads get? Matching parameter counts to within $\approx\!15\%$ makes the answer readable off the MAL numbers.

\section{Reproducibility details}
\label{app:reproducibility}
This section collects the setup details needed to reproduce our numbers. Code and weights are not released with this submission; the recipe below is complete enough for independent re-implementation.

\paragraph{Head architecture.} Each head is one transformer decoder layer at the backbone's width ($4096$ hidden, $32$ attention heads, $14336$ FFN, RMSNorm, SwiGLU, RoPE, matched to Qwen3-8B), followed by the frozen backbone LM head. Head $k$ receives $[h_t;\, \text{embed}(\arg\max_v p^{(k-1)}(v\mid h_t, \ldots))]$ as input; the concatenation is projected to hidden width by a linear layer before the transformer layer. Per-head parameter count is $\approx\!200$M ($\approx\!\text{one Qwen3-8B decoder layer}$), matched to MiMo-7B.

\paragraph{Training.} $5$ epochs on $\approx\!2.5$B target-generated tokens per epoch ($\approx\!19$k optimiser steps at effective batch $240$). AdamW with $\beta_1{=}0.9,\, \beta_2{=}0.95$, weight decay $0.1$, learning rate $5\times 10^{-5}$, cosine schedule to zero with $3\%$ linear warm-up. Gradient checkpointing on the frozen backbone (activations recomputed on the fly, no backbone parameters updated). Sequence length $2048$, packed. The $K{=}6$ variant uses the same schedule with two additional heads sharing per-head architecture.

\paragraph{On-policy corpus.} $1.89$M reasoning prompts from AM-Thinking~\citep{amthinking2024}. Each prompt is completed by the frozen Qwen3-8B in thinking mode with the deployment decoding parameters (temperature $0.6$, top-$p$ $0.95$, top-$k$ $20$). Prompt/response pairs are concatenated verbatim; no filtering.

\paragraph{Ablation architectures.} Each ablation replaces one component of the base recipe and is trained on the raw AM-Thinking corpus (Row~2 of Table~\ref{tab:ablations}) with the same schedule and optimiser as the base recipe.

\emph{GRU cross-head gate.} At each head boundary $k{-}1\!\to\!k$ we insert a single-layer GRU cell that consumes the current head's hidden state $h_t^{(k-1)}\in\mathbb{R}^{4096}$ and the argmax-token embedding of head $k{-}1$ (dimension $4096$, from the backbone embedding table), and emits a gated hidden state $\tilde h_t^{(k)}\in\mathbb{R}^{4096}$ that is consumed by head $k$ in place of the base $h_t$. Concretely, letting $g^{(k)}\!\in\!\mathbb{R}^{4096}$ be the GRU state, $g^{(k)} = \mathrm{GRU}(g^{(k-1)},\, [h_t^{(k-1)};\,\mathrm{embed}(\hat x_{t+k-1}^{(k-1)})])$ and $\tilde h_t^{(k)} = W_o\,g^{(k)} + h_t^{(k-1)}$ with $W_o\in\mathbb{R}^{4096\times 4096}$ and a residual connection to preserve the base signal. Gate parameters are $\approx\!50$M per handoff, trainable end-to-end; backbone frozen.

\emph{Mamba-style cross-head gate.} Same interface as the GRU gate but with a selective-state-space block~\citep{gu2023mamba} replacing the GRU cell. The block operates on a length-$K$ virtual sequence obtained by stacking $[h_t^{(0)}, \ldots, h_t^{(k-1)}]$ across head indices at position $t$, with $d_{\text{model}}=4096$, state dimension $d_{\text{state}}=16$, expansion factor $2$, and a length-$4$ convolution. Because our chain has only $K{=}3$ or $K{=}6$ steps, the selective scan reduces to a small unrolled recursion; we implement it as the reference formulation in \citet{gu2023mamba} without their kernel-fused variant, since the sequence length does not warrant the specialised CUDA path. The residual and output projection match the GRU gate. Gate parameters are $\approx\!80$M per handoff.

\emph{Shallow LoRA.} LoRA adapters~\citep{hu2021lora} of rank $r{=}8$, $\alpha{=}16$, dropout $0.05$, on the Q/K/V/O attention projections of the last $4$ of the $36$ backbone layers. Backbone learning rate $1\times 10^{-5}$ (heads at $5\times 10^{-5}$). Adapter parameter count $\approx\!1$M total.

\emph{Deep LoRA.} Same rank/$\alpha$/dropout/backbone LR as shallow, applied to the Q/K/V/O attention projections of the last $18$ of the $36$ backbone layers (i.e.\ half the backbone). Adapter parameter count $\approx\!4.7$M total.

\emph{Full-vocabulary KL.} Loss $D_{\text{KL}}(p_\theta \| p^{(k)})$ over the full vocabulary at every position, replacing per-token hard cross-entropy in Eq.~\ref{eq:loss}.

\paragraph{Verification.} vLLM V1 speculative decoding path with our verification-rule extensions in \S\ref{sec:approx_verify}. Chain threshold in Eq.~\ref{eq:joint} enforced in log-space with a clamp $\log(p)\geq -27.6$ (i.e.\ $p\geq 10^{-12}$).

\paragraph{Adaptive-$K$ controller (LCB).}
The controller is an online multi-armed bandit over six arms $\{K{=}1,\ldots,K{=}6\}$; Each arm's reward is the observed batched throughput $\mathrm{tps}_a = \text{accepted tokens}/\text{wall-clock}$ over a \emph{measurement window} of $W{=}8$ steps at near-full occupancy (a step counts only if the in-flight batch is $\geq 3/4$ of capacity, so wave-drain tails are dropped). After $\geq n_{\min}{=}2$ round-robin windows per arm, per-arm running estimates $(\hat\mu_a, \hat v_a)$ update by exponential smoothing with $\beta{=}0.7$:
\[
\hat\mu_a \leftarrow \beta\hat\mu_a + (1{-}\beta)\,\mathrm{tps}_a,
\qquad
\hat v_a \leftarrow \beta\hat v_a + (1{-}\beta)\,(\mathrm{tps}_a - \hat\mu_a^{\text{old}})^2,
\]
so $\hat s_a{=}\sqrt{\hat v_a}$ is the arm's measurement spread. The controller then exploits $\arg\max_a \hat\mu_a$ with periodic re-exploration every $P{=}8$ windows. \emph{Commit} fires the first time the argmax arm has been the argmax for $C{=}3$ consecutive post-settle windows \emph{and} leads the runner-up by a relative margin $\delta{=}0.015$ (i.e.\ $(\hat\mu_{(1)}-\hat\mu_{(2)})/\hat\mu_{(2)}\geq\delta$); if neither condition is met within $W_{\max}{=}30$ windows a commit is forced. At commit we select the arm with the highest lower-confidence bound $\hat\mu_a - z\hat s_a$ with $z{=}1$, rather than the raw argmax. After commit the depth is frozen for the rest of the session: no classifier forward, no window bookkeeping, no per-step host sync. Reported numbers correspond to the post-commit segment. A warm-up of $100$ steps precedes the first window to let the KV cache and CUDA-graph pool reach steady state.

\begin{algorithm}
\caption{Adaptive-$K$ throughput bandit (LCB , one serving session).}
\label{alg:bandit}
\begin{algorithmic}[1]
\State \textbf{Input:} arms $\mathcal A=\{K{=}1,\ldots,6\}$; window $W{=}8$; settle $n_{\min}{=}2$; explore period $P{=}8$; commit stability $C{=}3$; margin $\delta{=}0.015$; hard cap $W_{\max}{=}30$; LCB $z{=}1$; EMA $\beta{=}0.7$; occupancy $\geq 3/4$
\State $(\hat\mu_a,\hat v_a,N_a)\leftarrow(0,0,0)$ for $a\in\mathcal A$; $\text{best}_{\text{prev}}\!\leftarrow-1$; $\text{run}\!\leftarrow 0$; committed$\,\leftarrow-1$; $w\!\leftarrow 0$
\While{decoding}
  \State collect a window of $W$ near-full-occupancy steps under the current arm $a$; measure $\mathrm{tps}_a$
  \State $(\hat\mu_a, \hat v_a) \leftarrow (\beta\hat\mu_a+(1{-}\beta)\mathrm{tps}_a,\ \beta\hat v_a+(1{-}\beta)(\mathrm{tps}_a-\hat\mu_a^{\text{old}})^2)$; $N_a\!+\!\!=\!1$; $w\!+\!\!=\!1$
  \If{$\min_a N_a < n_{\min}$} \Comment{settle: round-robin to the least-sampled arm}
    \State $a \leftarrow \arg\min_a N_a$; \textbf{continue}
  \EndIf
  \State $\text{best}\leftarrow\arg\max_a \hat\mu_a$; $\text{run} \leftarrow \text{run}{+}1$ if $\text{best}{=}\text{best}_{\text{prev}}$ else $1$; $\text{best}_{\text{prev}}\leftarrow \text{best}$
  \State $\text{lead}\leftarrow (\hat\mu_{(1)}-\hat\mu_{(2)})/\hat\mu_{(2)}$
  \If{$\text{run}\geq C$ \textbf{and} $\text{lead}\geq\delta$, \textbf{or} $w\geq W_{\max}$}
    \State committed $\leftarrow \arg\max_a\ (\hat\mu_a - z\sqrt{\hat v_a})$ \Comment{LCB tie-break: reliably-good beats spiky}
    \State \textbf{return} committed \Comment{freeze depth for the rest of the session}
  \EndIf
  \State $a \leftarrow \text{best}$ if $w \bmod P \neq 0$ else next arm $\neq \text{best}$ \Comment{periodic re-exploration}
\EndWhile
\end{algorithmic}
\end{algorithm}
\paragraph{Evaluation.} We use the vLLM~\citep{leviathan2023spec} implementation of speculative decoding on Qwen3-8B for both our checkpoint and MiMo-7B-RL. Batch $4$, single A100-80GB GPU, thinking mode with the training-time decoding parameters. Math final answers extracted from \texttt{\textbackslash boxed\{$\cdot$\}} and scored by \texttt{math\_verify}; MCQ by single-letter match; code by unit-test execution; CNN/DailyMail~\citep{nallapati2016cnndm} by ROUGE-1/2/L~\citep{lin2004rouge} against reference highlights.

\paragraph{Compute.} Head training on $4$ nodes of $8\times$A100-80GB with tensor parallel $1$, data parallel $32$. One epoch on $2.5$B tokens takes $\approx\!8$h; $5$ epochs $\approx\!40$h. The $K{=}6$ variant costs $\approx\!\!1.4\times$ the $K{=}3$ variant. 

\section*{LLM use disclosure}
\label{app:llm}
Large language models were used for editing prose in this manuscript and for scaffolding experimental scripts and figure code. All numerical results, tables, figures, and technical claims were checked by the authors against source-code outputs; the LLM was not used to generate any of the reported numbers or to author any technical claim.

\end{document}